\documentclass[letterpaper, 10 pt, conference]{packages/ieeeconf}  

\IEEEoverridecommandlockouts                              

\usepackage{times}
\usepackage{epsfig}
\usepackage{graphicx}
\usepackage{amsmath}
\usepackage{amssymb}
\usepackage{gensymb}
\usepackage{subfigure} 
\usepackage[ruled]{algorithm}
\usepackage{algpseudocode}
\usepackage{ctable}
\usepackage[export]{adjustbox}
\usepackage{import}
\usepackage{pgf}
\usepackage{setspace}
\usepackage{url}
\usepackage{hyperref}
\usepackage{capt-of,etoolbox}

\usepackage{colortbl}
\makeatletter
\let\NAT@parse\undefined
\makeatother
\usepackage[numbers]{natbib}

\usepackage{packages/sparo_acronyms}
\usepackage{packages/sparo_math}
\let\celsius\relax

\let\degree\relax
\usepackage{packages/sparo_SIunits}
\usepackage{packages/sparo_misc}
\usepackage{multirow}
\usepackage{indentfirst}

\usepackage{soul,color}
\usepackage{verbatim} 

\usepackage{xcolor}

\usepackage[symbol]{footmisc}
\renewcommand{\thefootnote}{\fnsymbol{footnote}}

\renewcommand{\thefootnote}{\arabic{footnote}}

\begin{document}

\title{\LARGE \bf
   MARSCalib: Multi-robot, Automatic, Robust, Spherical Target-based Extrinsic Calibration in Field and Extraterrestrial Environments

}

\author{Seokhwan Jeong$^{1}$, Hogyun Kim$^{1}$, and Younggun Cho$^{1\dagger}$
	\thanks{This work was supported by Institute of Information \& communications Technology Planning \& Evaluation (IITP) grant (RS-2022-II220448), National Research Foundation of Korea (NRF) grant (RS-2025-02217000) funded by the Korea government (MSIT) and  Smart Manufacturing Innovation R\&D funded by Korea Ministry of SMEs and Startups (RS-2024-00448642).}
	\thanks{$^{1}$Seokhwan Jeong, $^{1}$Hogyun Kim and $^{1\dagger}$Younggun Cho are with the Electrical and Computer Engineering, Inha University, Incheon, South Korea 
		{\tt\small [eric5709, hg.kim]@inha.edu, yg.cho@inha.ac.kr}}%
}

\maketitle

\begin{abstract} 

This paper presents a novel spherical target-based LiDAR-camera extrinsic calibration method designed for outdoor environments with multi-robot systems, considering both target and sensor corruption. The method extracts the 2D ellipse center from the image and the 3D sphere center from the pointcloud, which are then paired to compute the transformation matrix. Specifically, the image is first decomposed using the Segment Anything Model (SAM). Then, a novel algorithm extracts an ellipse from a potentially corrupted sphere, and the extracted ellipse’s center is corrected for errors caused by the perspective projection model. For the LiDAR pointcloud, points on the sphere tend to be highly noisy due to the absence of flat regions. To accurately extract the sphere from these noisy measurements, we apply a hierarchical weighted sum to the accumulated pointcloud. Through experiments, we demonstrated that the sphere can be robustly detected even under both types of corruption, outperforming other targets. We evaluated our method using three different types of LiDARs (spinning, solid-state, and non-repetitive) with cameras positioned in three different locations. Furthermore, we validated the robustness of our method to target corruption by experimenting with spheres subjected to various types of degradation. These experiments were conducted in both a planetary test and a field environment. Our code is available at https://github.com/sparolab/MARSCalib.

\end{abstract}


\section{Introduction}
LiDAR-camera sensor fusion techniques have moved beyond the research stage and are now widely applied in real-world applications, including autonomous driving \cite{yeong2021sensor}, \ac{SLAM} \cite{xu2022review}, and other domains \cite{tran2023adaptive}. For effective fusion, extrinsic calibration must accurately determine the spatial relationship between each sensor's coordinate system.

However, these extrinsic parameters are not permanent. Consider a multi-modal sensor platform in which calibration is performed prior to mission deployment. During operation, sensors may deviate from their initial poses due to temperature fluctuations or harsh vibrations \cite{jing2023online}. Even minor drift can significantly degrade system performance and potentially lead to mission failure. Therefore, extrinsic calibration must be repeated whenever drift is detected, ensuring system safety and accuracy.

\begin{figure}[t]
    \centering
	\def\width{0.98\columnwidth}
	\includegraphics[width=\width, trim=0 0 0 0, clip]{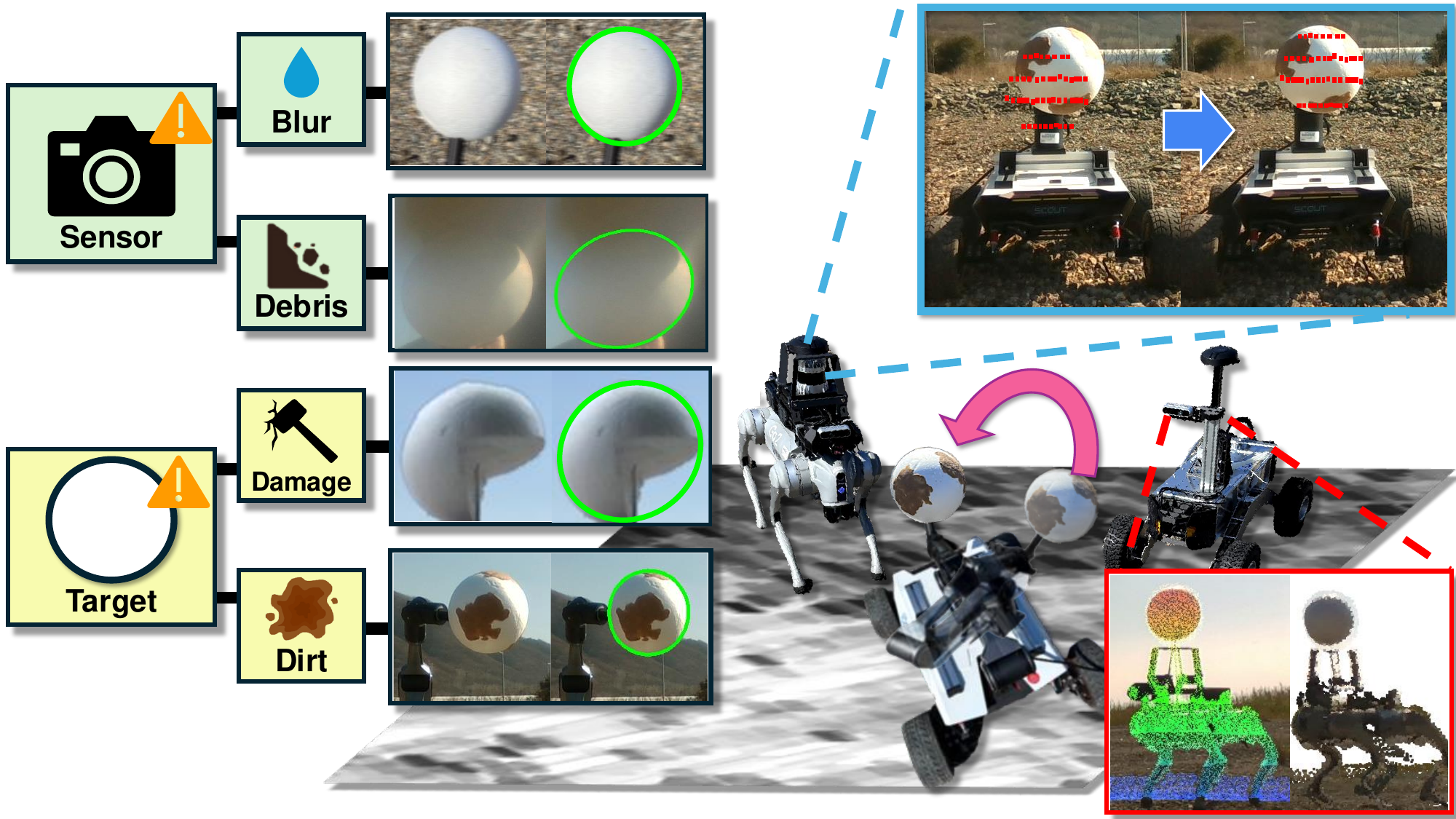}
    \caption{Illustration of our extrinsic calibration approach for a multi-robot system in a planetary test environment, showing both sensor- and target-level corruption. \textbf{Left:} Examples of potential corruption types (blur, debris, damage, dirt) and corresponding ellipse-detection results under outdoor conditions. \textbf{Right:} Quantitative calibration result using a spherical target. The quadruped robot (blue box) uses a spinning LiDAR and a camera, while the wheeled rover (red box) uses a non-repetitive LiDAR and a camera.}
    \label{figs:thumbnail}
    \vspace{-0.4cm}
\end{figure}

In this work, we consider two scenarios for performing extrinsic calibration in multi-robot deployments such as planetary \cite{varadharajan2025multi, kim2025skid} and radiation-hazardous \cite{ardiny2024enhancing} exploration sites, where human intervention is limited, as well as automated farms~\cite{ju2022review}, where human involvement should be minimal. In the first scenario, sensor drift is detected during operation, prompting another robot equipped with a target for recalibration. In the second scenario, a robot starts operation for the first time or undergoes sensor replacement due to mission modifications.


From a perspective of robot employing calibration, target-based methods that utilize checkerboards \cite{huang2024novel, geiger2012automatic}, polygonal planes \cite{park2014calibration, liao2018extrinsic}, or perforated planes \cite{zhang20232, beltran2022automatic} are often impractical. They tend to be large, sensitive to viewpoint changes, and are difficult for robots to carry. In addition, outdoor environments—unlike controlled laboratory settings—feature mud, dirt, and rocks that may contaminate or damage both the sensor and the target. For instance, a checkerboard loses accuracy if any corner is occluded or worn. Furthermore, in the context of sensor corruption, methods relying on shape-based characteristics generally handle noisy or blurry data more robustly than those depending on geometric features such as lines or corners. Consequently, a spherical target is well suited to our proposed scenarios because it is invariant to viewing direction, feasible for robots to transport, and remains reliable under both target and sensor corruption.

In this paper, we propose a novel LiDAR-camera extrinsic calibration method designed for outdoor environments and multi-robot systems, accounting for potential corruption in both the spherical target and the sensors as shown in \figref{figs:thumbnail}. The main contributions of this paper are as follows:

\begin{itemize}
    \item \textbf{M}ulti-robot calibration: We propose a calibration method for multi-robot systems, enabling the completion of the calibration procedure without human intervention.
    \item \textbf{A}utomatic detection: Our system automatically detects the sphere center in both LiDAR and camera data, thus ensuring a fully autonomous calibration process.
    \item \textbf{R}obust detection: In scenarios where the target is prone to damage and replacements are limited, our method reliably extracts the sphere’s geometry even under target and sensor degradation. To the best of our knowledge, this is the first study comparing the robustness of a spherical target with other conventional targets in corrupted settings.
    \item \textbf{S}pherical target: We employ a spherical target that is effortless for robots to carry in outdoor environments. Also, this work employs the smallest spherical target to date for LiDAR-camera calibration.
\end{itemize}

\section{Related Works}

The core challenge in LiDAR–camera extrinsic calibration lies in bridging the gap between their disparate dimensional representations of the shared environment. Approaches can be broadly categorized into \textit{target-based} and \textit{targetless} methods depending on how calibration information is extracted. Target-based methods rely on the geometric features of predefined targets and necessitate human intervention, whereas targetless methods are further divided into (i) scene-based methods that exploit environmental cues, (ii) motion-based methods relying on sensor movement, and (iii) deep-learning methods leveraging data-driven feature extraction. Finally, we discuss \textit{target and sensor corruption}, an often-overlooked problem that is critical in real-world deployments.

\begin{figure*}[t]
    \centering
	\def\width{0.9\textwidth}%
	\includegraphics[width=\width, trim=0 0 0 0, clip]{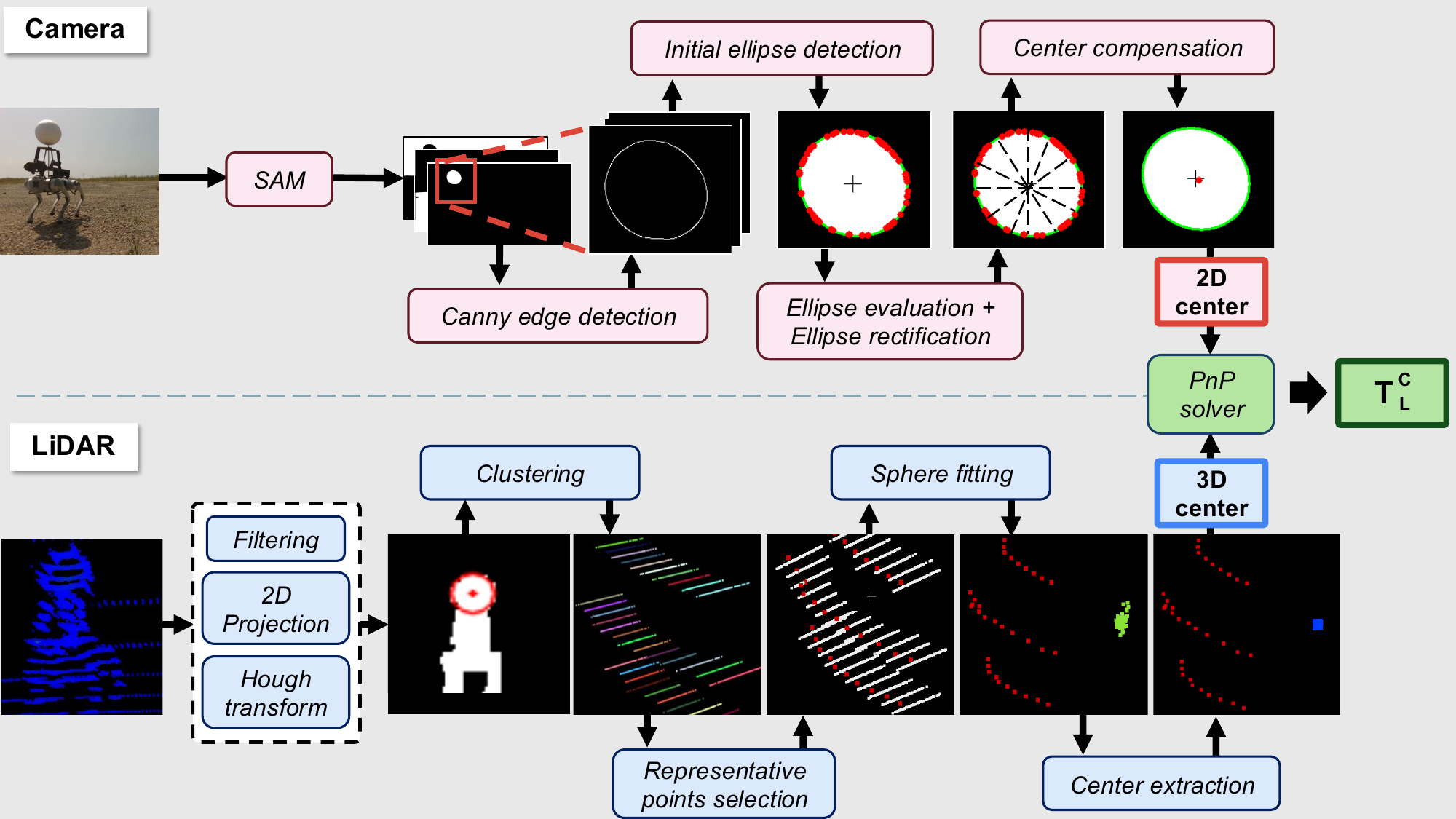}
 \caption{Overview of the proposed method. 
The camera pipeline (top) detects the ellipse and extracts its 2D center, while the LiDAR pipeline (bottom) estimates the sphere's 3D center. 
These center pairs are then fed into a PnP solver to compute the extrinsic parameters $\mathbf{T}^C_L$.}
    \label{figs:pipeline}
    \vspace{-0.4cm}
\end{figure*}

\subsection{Target-based approaches}
\citet{geiger2012automatic} introduced a single-shot calibration method using multiple checkerboards, while \citet{wang2017reflectance} refined it by estimating 3D corners from pointcloud through intensity data. \citet{xie2018pixels} enhanced checkerboard methods by incorporating square openings for direct 3D-2D alignment, thereby improving robustness. Despite their popularity, however, checkerboard-based approaches can be cumbersome in multi-robot or outdoor settings due to size constraints and vulnerability to damage or occlusion. Planar boards represent another widely studied class. \citet{park2014calibration} proposed triangular and diamond-shaped boards, whereas \citet{beltran2022automatic} designed boards with circular holes. Nonetheless, these methods are sensitive to wind interference and difficult to manufacture, and the latter also requires a flat backdrop, limiting their usage in complex outdoor terrains. Spherical targets \cite{toth2020automatic, zhang2024automatic,chen2018extrinsic,kummerle2018automatic} offer a geometry that is inherently invariant to viewing direction, simplifying alignment. However, they use a large sphere with a radius of at least 25 cm, assume relatively noise-free LiDAR data, and require a clearly visible sphere in camera images. In practice, pointcloud measurements often contain considerable noise, and partial occlusion or small sphere size can make ellipse detection less reliable. Moreover, previous methods rarely discuss \textit{target corruption} (e.g., dents, scratches, or partial contamination), which can distort the sphere's shape.

To address these challenges, we propose a hierarchical weighted-sum approach that robustly extracts the sphere from a noisy pointcloud and a segmentation strategy based on the Segment Anything Model (SAM)~\cite{kirillov2023segment} for detecting small or partially corrupted targets in images.

\subsection{Targetless approaches}

Scene-based methods, such as \citet{yuan2021pixel}, rely on depth-discontinuous edges to perform calibration while accounting for LiDAR noise. Similarly, \citet{koide2023general} introduced a general extrinsic calibration toolbox leveraging learning-based features. However, in low-texture environments, such scene-based methods can struggle to identify distinctive cues, reducing their accuracy. Motion-based methods exploit sensor movement. For instance, \citet{ishikawa2018lidar} used a camera motion scale to solve the hand-eye calibration problem, but its accuracy heavily depends on the quality of each sensor's pose estimation. Deep-learning approaches like RegNet \cite{schneider2017regnet} and LCCNet \cite{lv2021lccnet} apply \acp{CNN} to achieve multi-modal calibration with near real-time capability, but they still struggle to generalize to new sensor setups and environments.

\subsection{Target \& Sensor Corruption in Field Environments}

In real-world field scenarios, both calibration targets and sensors can be corrupted by (i) \textit{contamination}, including mud, dirt, or dust that obscures targets or lenses, and (ii) \textit{physical damage}, such as cracks or scratches that distort geometry. In the case of target corruption, \citet{claro2023artuga}, \citet{fiala2005artag}, and \citet{yahya2017detection} examined fiducial marker failures under occlusion in \ac{UAV} landing systems, spacecraft docking, and \ac{AUV} docking, respectively. \citet{bian2024coppertag} proposed a fiducial marker that can be robustly detected despite rotation and occlusion, while \citet{chen2021automatic} developed checkerboard-based camera calibration against partial occlusion. In contrast, for sensor corruption, \citet{uricar2021let} and \citet{zhang2023perception} tackled soiled camera lenses that reduce image clarity.

Despite this progress, existing studies consider only one type of corruption-either the target or the sensor-but not both. Moreover, calibration research still lacks investigations that address these degradations. Motivated by this gap, our work aims to preserve calibration accuracy even under such adverse conditions.

\begin{figure}[!t]
    \centering
	\def\width{0.98\columnwidth}%
	\includegraphics[width=\width, trim=1.18cm 0cm 6cm 0cm, clip]{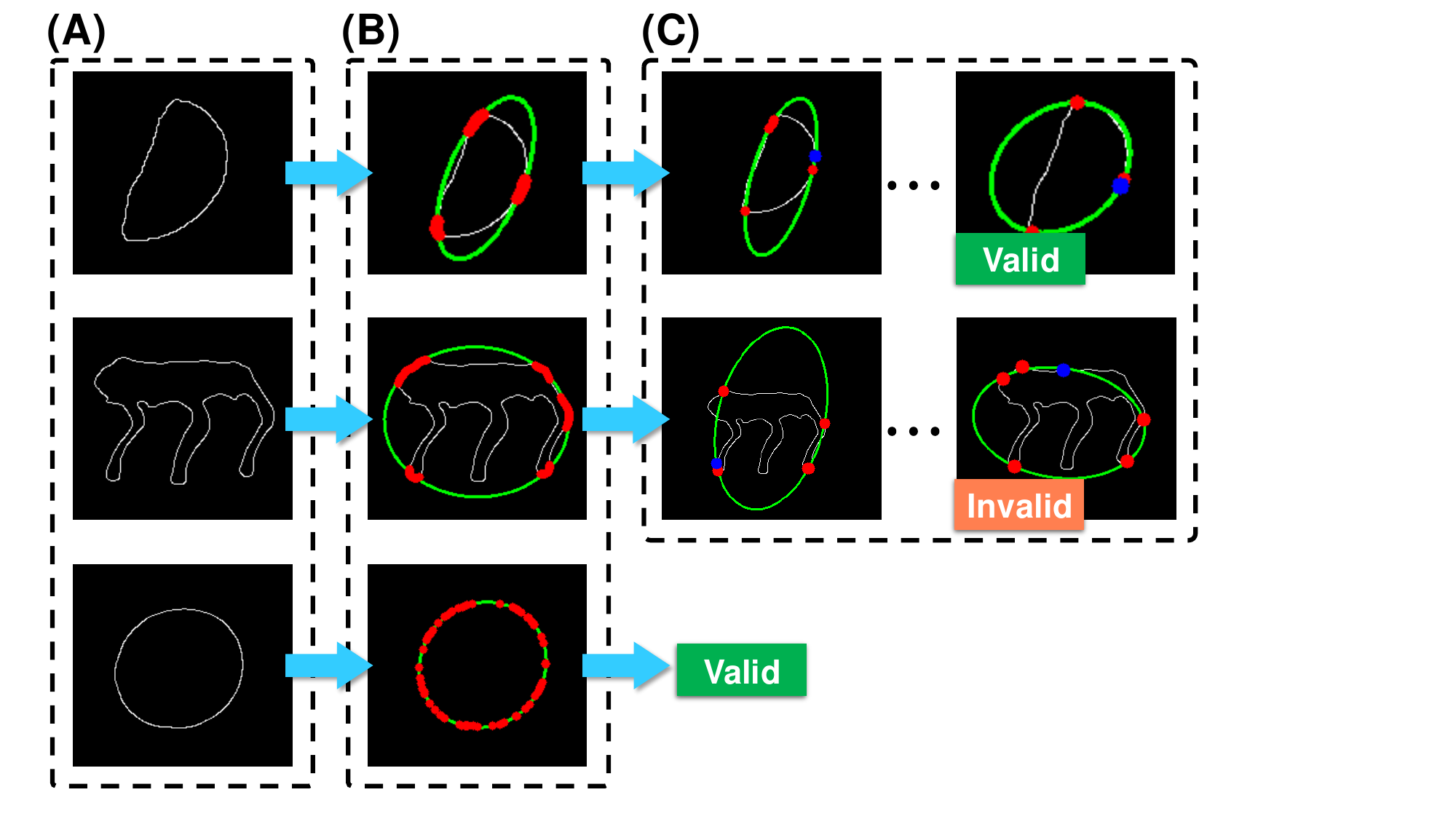}
    \vspace{-0.4cm}
     \caption{
     Illustration of the process from \emph{Initial ellipse detection} to \emph{Ellipse rectification}. (A) shows edge images of a severely damaged or contaminated sphere (first row), a non-spherical object (second row), and an undamaged sphere (third row). (B) displays the output of \emph{Initial ellipse detection}, which then proceeds to \emph{Ellipse evaluation}. The sampled points ($\mathcal{P}_{e}$, shown in red) in the first and second rows are concentrated in specific regions, leading to the \emph{Ellipse rectification}. In contrast, the ellipse in the third row is classified as valid because $\mathcal{P}_{e}$ is evenly distributed. (C) represents the \emph{Ellipse rectification}. The ellipse is rectified by iteratively fitting with a point (blue) that does not belong to $\mathcal{P}_{e}$, along with points from each concentrated region (red), which represent intact regions. As a result, ellipses from the first and third rows are determined to be valid and proceed to the \emph{Center compensation}.     
     }
    \label{figs:ellipse_detection}
    \vspace{-0.6cm}
\end{figure}


\section{Method}
\figref{figs:pipeline} illustrates the pipeline of our proposed method. The process is summarized as follows. In the camera pipeline, we apply the Segment Anything Model (SAM) before performing Canny edge detection \cite{canny1986computational}, which enhances the detection of small and potentially corrupted spheres. After edge detection, we extract an ellipse and evaluate it through the following steps: \emph{Initial ellipse detection}, \emph{Ellipse evaluation}, and \emph{Ellipse rectification}. We then perform \emph{Center Compensation} to compensate for error arising from the extracted ellipse. 

For LiDAR data, we first identify points in the sphere region. Next, we use a hierarchical weighted-sum approach to both estimate the sphere and derive the corresponding center of the sphere. Finally, we estimate the extrinsic parameters by minimizing the reprojection error between the derived 3D-2D center pairs, using the Levenberg-Marquardt optimizer~\cite{ranganathan2004levenberg}.

\subsection{Ellipse Center Extraction (Camera)}
\textbf{Preprocessing} $-$ First, we segment the image using SAM~\cite{kirillov2023segment}, which produces a set of mask images. As SAM narrows down the search region, it facilitates the detection of smaller spheres. We then apply Canny edge detection to each mask image.

\textbf{Initial Ellipse Detection} $-$ This step examines the presence of an ellipse by sampling points from the edge images. All points on the edges used for sampling are denoted as $\mathcal{P}_{s}$. If points from impaired regions are included in the sample for ellipse fitting, they lie inside the true ellipse, causing the fitted ellipse to exhibit greater eccentricity and appear more elongated than the true ellipse. Considering this property, iteratively excluding the points inside the detected ellipse from $\mathcal{P}_{s}$ ensures that the remaining points belong to intact regions. After fitting an ellipse that satisfies the aforementioned condition, the sampled points that constitute the ellipse are denoted as $\mathcal{P}_{e}$.
If $\mathcal{P}_{s}$ is exhausted before an ellipse is found, the edge image is considered invalid.

\textbf{Ellipse Evaluation} $-$ To decide whether the initial ellipse derives from an intact spherical target, we build an angle-based histogram of the sampled points $\mathcal{P}_{e}$. If $\mathcal{P}_{e}$ is evenly distributed along the ellipse perimeter, the ellipse is deemed valid; otherwise, it is classified as originating from a non-spherical object or a damaged target and is forwarded to the next step.

\textbf{Ellipse Rectification} $-$  Ellipse not derived from an intact spherical target can be biased in certain cases since $\mathcal{P}_{e}$ clusters in specific regions. The bias grows as the degree of corruption increases. To mitigate this effect, we iteratively re-fit the ellipse using two types of samples: single point randomly drawn from $\mathcal{P}_{s} \cap \mathcal{P}_{e}^{\mathsf{c}}$
and points from each concentration region. If a certain portion of $\mathcal{P}_{s}$  lies outside the re-fitted ellipse, the ellipse is considered invalid; otherwise, it is deemed valid and regarded as originating from a corrupted sphere. An example of the process from \emph{Initial ellipse detection} to  \emph{Ellipse rectification} is provided in \figref{figs:ellipse_detection}.

\begin{figure}[h]
    \centering
	\def\width{0.98\columnwidth}%
	\includegraphics[width=\width, trim=0 0 0 0, clip]{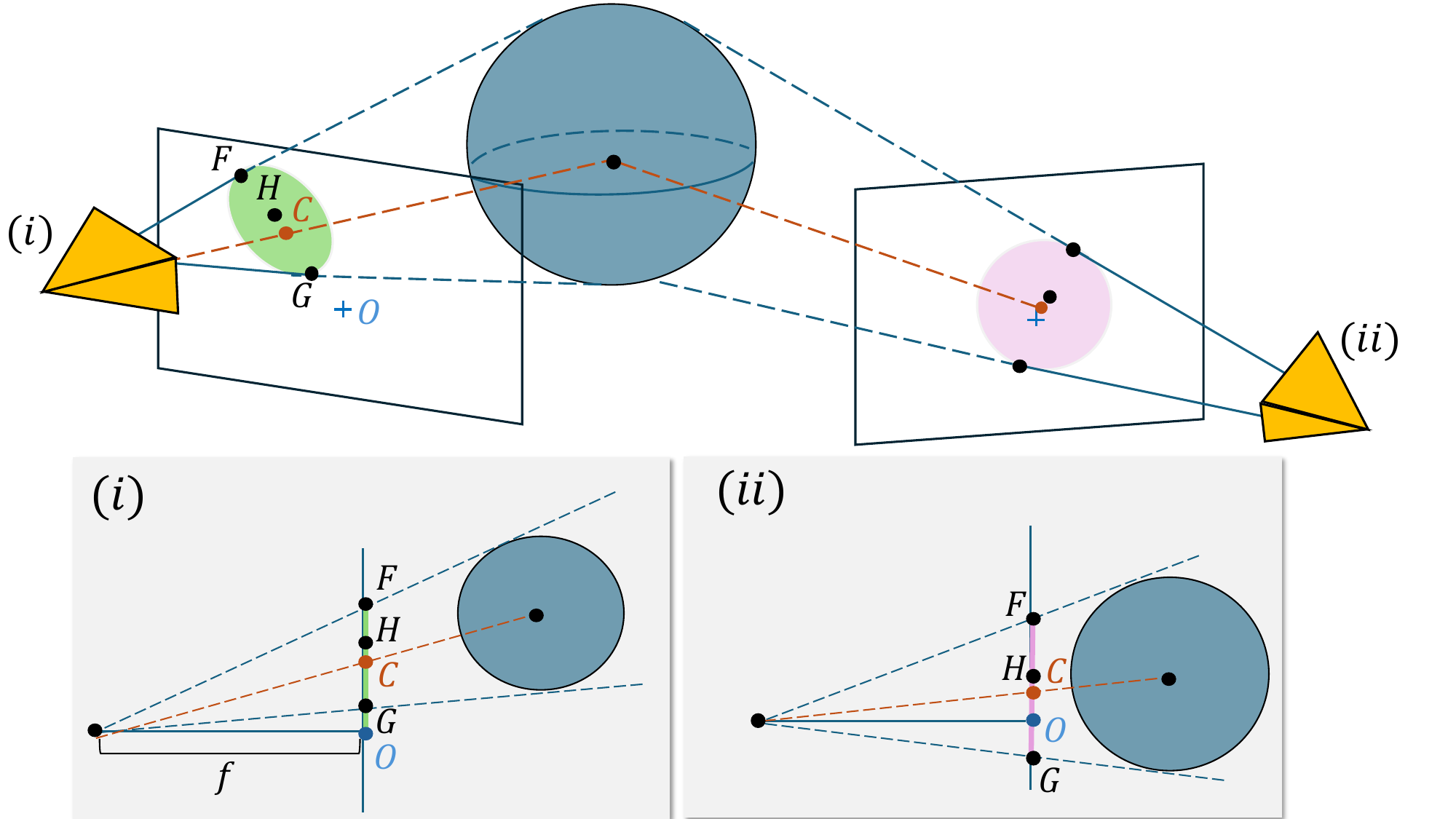}
 \caption{Perspective distortion of a spherical target. 
Compensation can be categorized into two cases: (i) when the image center is located outside the ellipse, and (ii) when it lies inside the ellipse.}
    \label{figs:projection}
\end{figure}

\textbf{Center Compensation} $-$ When a sphere is projected onto the image plane, it is represented as an ellipse unless the center of the sphere aligns with the very center of the image plane as shown in \figref{figs:projection}, leading to an error \cite{toth2023minimal, zhang2024automatic}. Compensation for the error $\epsilon$ is achieved by calculating the discrepancy between the projected ellipse center $H$ and the true center $C$. This process is divided into two cases: (i) when the center of the image plane is outside the ellipse, and (ii) when it is inside the ellipse. The error $\epsilon$ is calculated as:
\begin{equation}
    \epsilon = \begin{cases} 
    \overline{OG} + \overline{HG} - \overline{OC}, & \text{if (i)} \\
    \overline{OH} - \overline{OC}, & \text{if (ii)}
    \end{cases}.
\end{equation}
Here, $O$ represents the center of the image plane, and $\overline{HG}$ denotes half of the major axis length.
The distance $\overline{OC}$ is computed as follows:
\begin{equation}
\small
    \overline{OC} = \begin{cases}
    f \cdot \tan\left(\frac{1}
    {2}\left[\arctan\left(\frac{\overline{OF}}{f}\right) + \arctan\left(\frac{\overline{OG}}{f}\right)\right]\right), & \text{if (i)} \\
    f \cdot \tan\left(\frac{1}{2}\left[\arctan\left(\frac{\overline{OF}}{f}\right) - \arctan\left(\frac{\overline{OG}}{f}\right)\right]\right), & \text{if (ii)}
    \end{cases}.
\end{equation}
Here, $f$ represents the focal length of the camera, which is obtained by the camera’s intrinsic calibration. Finally, the center of the projected ellipse is adjusted toward $O$ by the magnitude of the error $\epsilon$ as depicted in \figref{figs:pipeline}, at the red point in the result image of  \emph{Center Compensation}.

\subsection{Sphere Center Extraction (LiDAR)}
\textbf{Preprocessing} $-$ Unlike existing methods that utilize \ac{RANSAC} to decompose a spherical target within the entire scene, our approach employs the Hough transform \cite{hough1962method} on the projected image to identify the relatively small target region. The process is as follows: First, the pointcloud is filtered by applying Statistical Outlier Removal (SOR) \cite{rusu20113d} and ground segmentation to retain the robot and the target. Subsequently, the decomposed region is projected onto an image, and the sphere region is segmented using the Hough transform.

\begin{figure}[!t]
    \centering
	\def\width{\columnwidth}%
	\includegraphics[width=0.98\width, trim=5cm 1.1cm 5cm 2.5cm, clip]{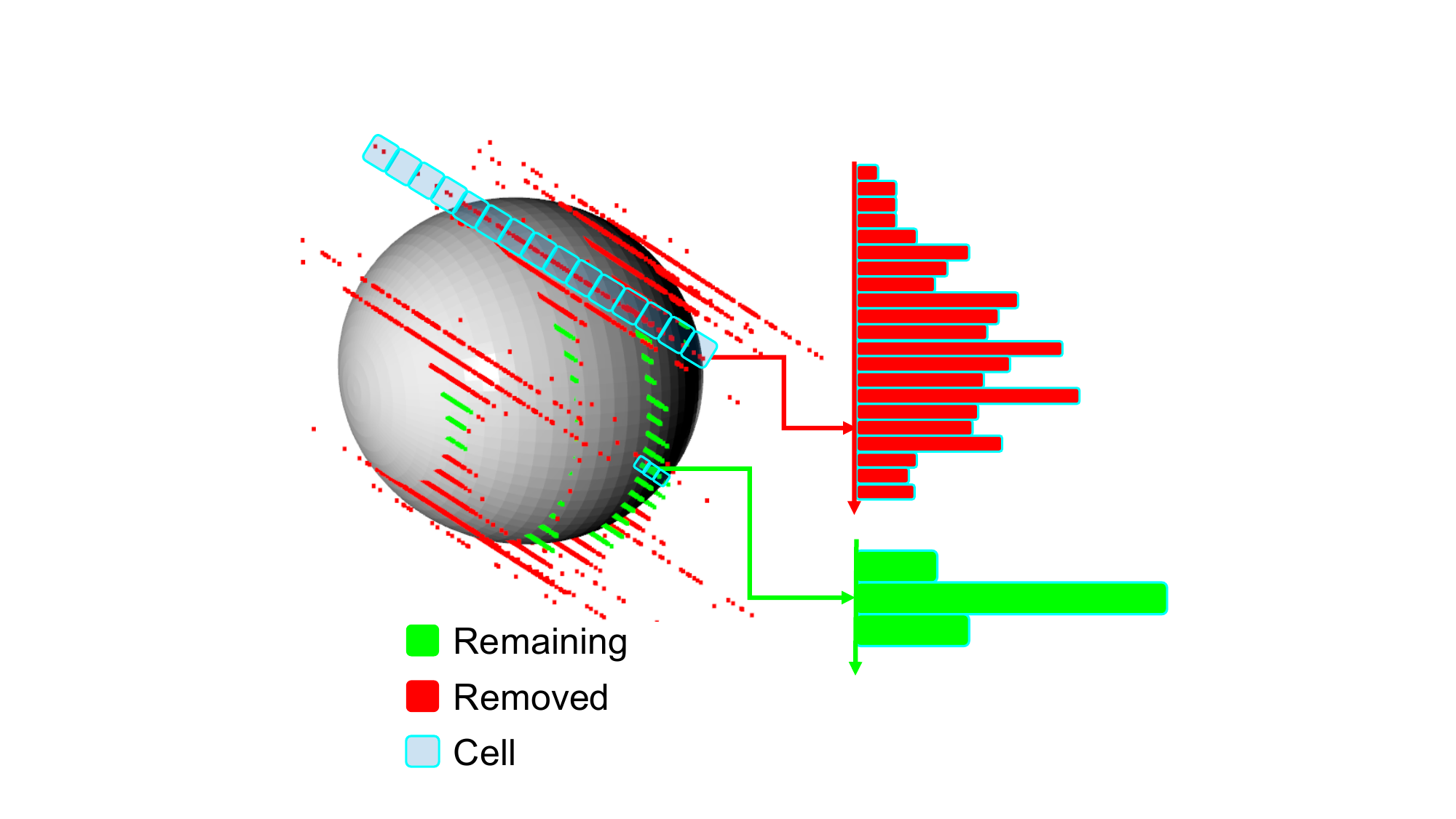}
 \caption{From the pointcloud, each line indicates a cluster. Red clusters, which indicate noisy clusters caused by large variance, are removed. Green clusters have relatively low noise. The graph shows the distribution of points in both removed and remaining clusters.}
    \label{figs:lidar_fig}
  \vspace{-0.4cm}
\end{figure}

\textbf{Clustering \&\ Representative Points Selection} $-$ We assume that the LiDAR measurements follow a probabilistic model with zero-mean Gaussian noise, which is influenced by both distance and incidence angle \cite{laconte2019lidar}. For spherical objects, the error increases as the laser scans reach further from the center due to the growing incidence angle. Therefore, estimating a sphere without accounting for noise is unreliable. To address this issue, we accumulate LiDAR data over time while keeping both the target and the LiDAR stationary. For spinning and solid-state LiDARs, points reflected from the same location on the sphere's surface form a cluster due to noise, aligning along the ray direction. The appearance of these clusters is depicted in Fig.\ref{figs:lidar_fig}. Clusters with greater lengths are removed, as they indicate the presence of significant noise. We subdivide each remaining cluster into equally sized cells and apply a weighted sum based on frequency to select a representative point ($p_r$) for each cluster, grouping them into~$\mathcal{P}_{r}$:
\begin{equation}
   p_r = \frac{\sum_{i=1}^M n_i \cdot c_i}{M},
\end{equation}
where $M$ denotes the total number of cells in the cluster, $n_i$ refers to the number of points in $i$-th cell, and $c_i$ denotes the location of $i$-th cell. For non-repetitive LiDAR, where the accumulated data does not exhibit a ray-like pattern, we voxel the data to extract representative points.

\textbf{Sphere Fitting $\&$ Center Extraction} $-$ Subsequently, to fit a sphere, we select all possible combinations of four points from $\mathcal{P}_{r}$, i.e.\ $\binom{\mathcal{P}_{r}}{4}$. Among these, combinations that form a plane are excluded, as they cannot define a sphere. The remaining combinations are used to parameterize spheres by solving a linear least-squares problem. The equation of the sphere is given by:

\begin{equation}
    x^2 + y^2 + z^2 = Ax + By + Cz + D,
\end{equation}
where $A$, $B$, $C$, and $D$ are the unknown parameters. This equation can be rewritten in matrix form as:

\begin{equation}
    \overrightarrow f = \mathbf{X}\cdot \overrightarrow p,
\end{equation}
where
\begin{equation}
\begin{bmatrix}
x_1^2 + y_1^2 + z_1^2 \\
x_2^2 + y_2^2 + z_2^2 \\
\vdots \\
x_j^2 + y_j^2 + z_j^2 
\end{bmatrix}
=
\begin{bmatrix}
x_1 & y_1 & z_1 & 1 \\
x_2 & y_2 & z_2 & 1 \\
\vdots & \vdots & \vdots & \vdots \\
x_j & y_j & z_j & 1
\end{bmatrix}
\begin{bmatrix}
A \\
B \\
C \\
D
\end{bmatrix}.
\end{equation}
The parameters $A$, $B$, $C$, and $D$ are then computed by solving the least-squares problem:
\begin{equation}
    \overrightarrow p = (\mathbf{X}^T \mathbf{X})^{-1} \mathbf{X}^T \overrightarrow f,
\end{equation}
where $\overrightarrow p$ represents the vector of unknown parameters. The resulting spheres are evaluated based on the known radius, and the optimal center point is determined by applying a weighted sum of the evaluated centers according to their respective frequencies. The evaluated centers and the optimal center are illustrated in \figref{figs:pipeline} as green and blue points, respectively. The supplementary information of the test is provided at our project page\footnotemark{}.

\subsection{Optimization}
After estimating the centers in both sensors, the final step is to determine the LiDAR-camera transformation matrix, $\mathbf{T}^C_L$. This is achieved by solving the Perspective-n-Point (PnP) problem, which minimizes the reprojection errors between the 3D points from the LiDAR and the corresponding 2D points from the camera. The Levenberg-Marquardt optimization method is employed to estimate the relationship.
The transformation matrix $\mathbf{T}^C_L$ is computed as:

\begin{equation}
    \mathbf{T}^C_L = \arg \min \sum_{k=1}^N \rho(\lVert \pi(\mathbf{T}^C_L p_k^{L}) - p_k^{C} \rVert), 
\end{equation}

where $p_k^{L}$ represents the $k$-th 3D point in the LiDAR coordinate system, $p_k^{C}$ denotes the corresponding 2D point in the camera image, and $\pi$ is the projection function.
To reduce the influence of outliers during optimization, we apply the robust kernel $\rho$.

 As non-linear optimization cannot completely remove the bias from inaccurate inputs, false detections of non-target objects (e.g., wheels and robot joints) are eliminated based on a threshold, followed by re-computation of the results.

\section{Experimental Results}

In this section, we first describe the outdoor environments and sensor configurations, and then present comprehensive experimental results that compare our proposed approach with existing methods under various scenarios. We focus on three key aspects: a) target detection comparison under challenging scenarios, b) calibration performance on various LiDAR-camera combinations, and c) calibration robustness to the damaged and contaminated targets. Additional materials can be found in the supplement video\footnotemark{}{\def\thefootnote{1,2}\footnotetext{https://sparolab.github.io/research/marscalib/}}.

\begin{figure}[t]
    \centering
    \def\width{0.98\columnwidth}%
    \includegraphics[width=\width, trim=0 0 10.3cm 0, clip]{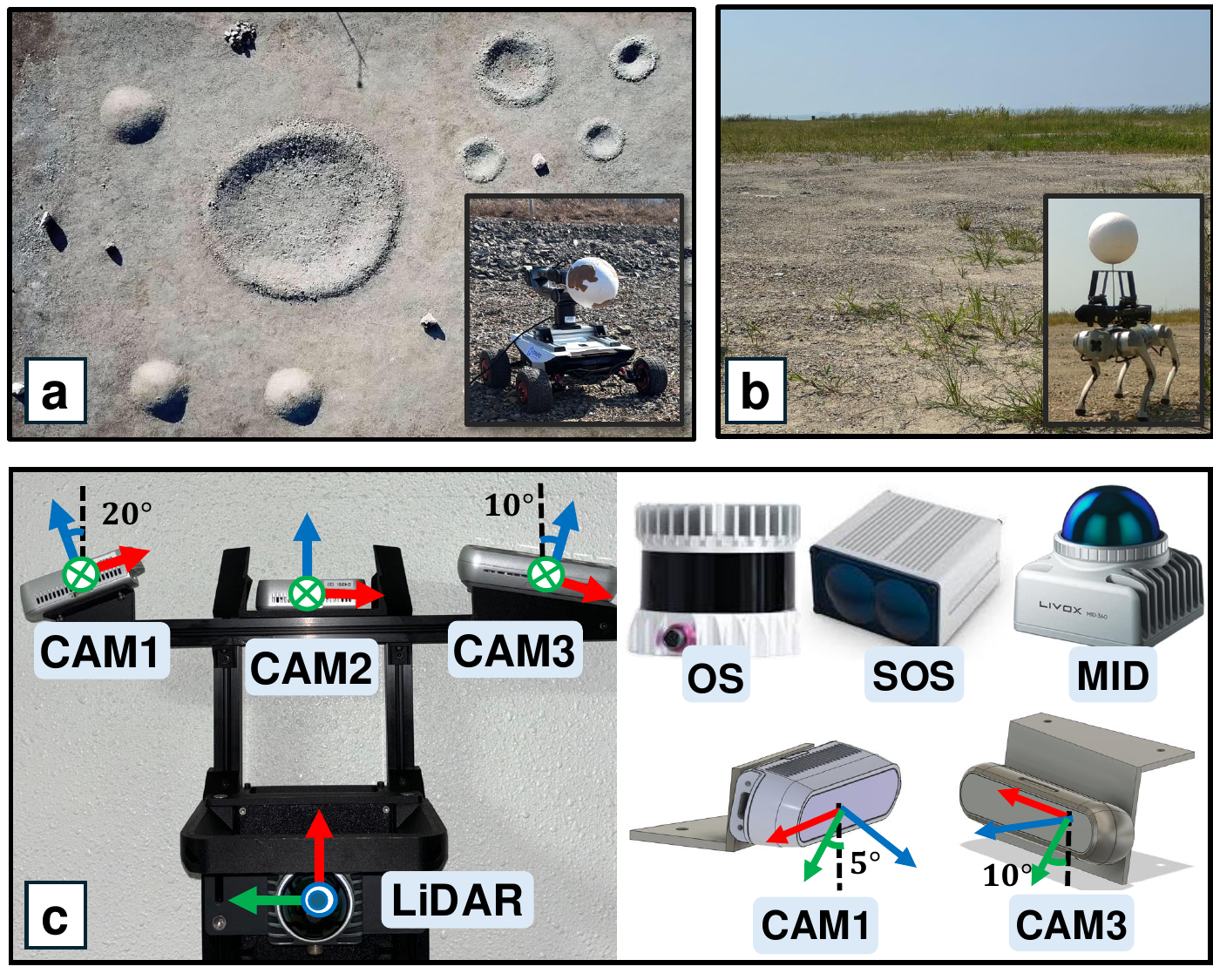}
    \vspace{-0.2cm}
     \caption{Test environments and setups. (a) A rover with a manipulator carrying a spherical target in a simulated planetary environment. (b) A quadrupedal robot with a spherical target mounted in a construction field. (c) Illustration of sensor mount for calibration. We design LiDAR-centric sensor mounts with cameras with different viewing angles.}
    \label{figs:environments}
    \vspace{-0.4cm}
\end{figure}



\subsection{Test Environments and Sensor Configurations}

As shown in \figref{figs:environments}(a) and \figref{figs:environments}(b), we evaluated our calibration approach in two challenging outdoor settings: a \emph{simulated planetary environment} designed to mimic moon terrains and a \emph{construction field} with uneven ground. For the spherical target, we used a readily available styrofoam sphere ($10\cm$ radius).


For the thorough evaluation, we used custom sensor setups equipped with three cameras (Realsense D435i (\textsc{Cam1} \& \textsc{Cam2}) and D455 (\textsc{Cam3})) with different viewing angles and three LiDAR types, each with its own scanning pattern: \textsc{Ouster OS1-32 (OS)}, \textsc{SOSLAB ML-X 120 (SOS)} and \textsc{Livox MID 360 (MID)} as represented in  \figref{figs:environments}(c). 


To establish a ground-truth reference, we used MATLAB’s LiDAR-camera extrinsic calibration toolbox \cite{zhou2018automatic} with a large-size checkerboard ($1200\mm \times 850\mm$).  We also specifically chose images and LiDAR frames for accurate calibration.

\subsection{Target Detection under sensor \& target contamination}

For target contamination, we evaluated the robustness of the spherical target against different viewing angles, comparing it to two fiducial markers: AprilTag \cite{olson2011apriltag}, a widely used marker, and CopperTag \cite{bian2024coppertag}, which is designed for occlusion resilience.

\begin{figure}[t!]
    \centering
	\def\width{0.95\columnwidth}%
	\includegraphics[width=\width, trim=0cm 8cm 15cm 0 , clip]{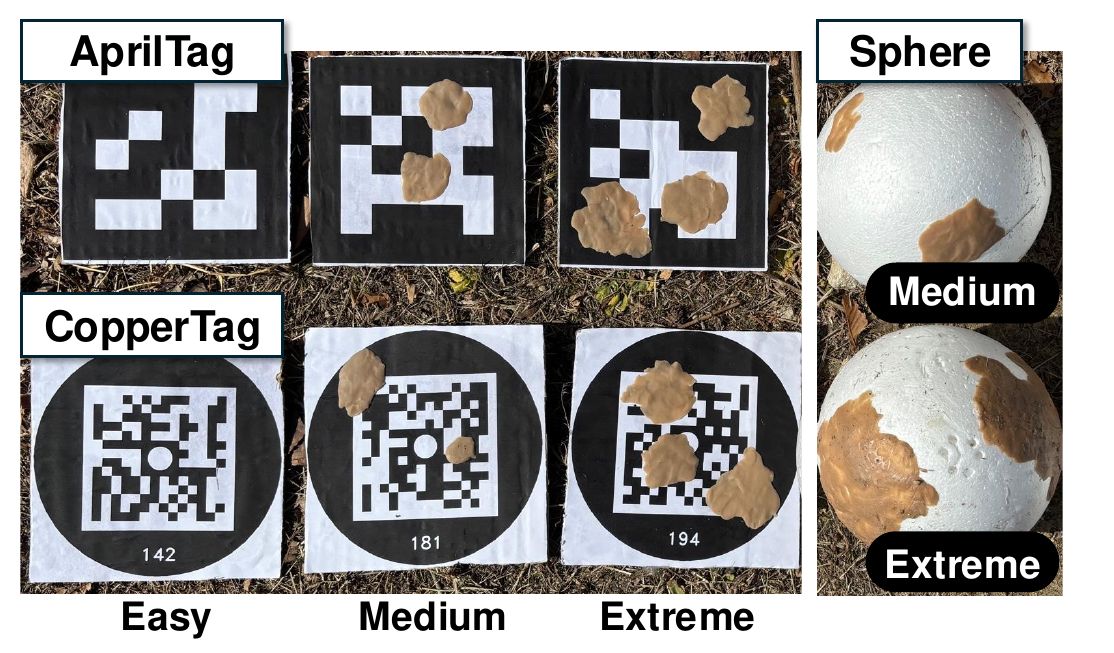}
    \vspace{-0.2cm}
 \caption{Examples of AprilTag, CopperTag, and a sphere target with different contamination levels.}
    \label{figs:cont_targets}
\end{figure}

We applied three levels of contamination (easy, medium, extreme) to three separate targets, as shown in \figref{figs:cont_targets}.
Each target was initially aligned with the sensor, then rotated from \(0^\circ\) to \(60^\circ\) in \(15^\circ\) increments around its ground-normal axis, both clockwise and counterclockwise.
If a marker or the sphere was detected in both directions, we marked it with a circle; if detected only once, with a triangle; if not detected, with a cross as indicated in \tabref{tab:detection}.


For sensor contamination, we tested two conditions: (i) using a semitransparent diffuser to blur the camera and (ii) obscuring the lens with mud.
We divided the image into six regions and placed the target in each region without rotation, then evaluated detection based on the total number of successful detections across these six placements. 

\begin{figure}[t]
    \centering
    \def\width{0.93\columnwidth}%
    \includegraphics[width=\width, trim=2cm 0 4.0cm 0, clip]{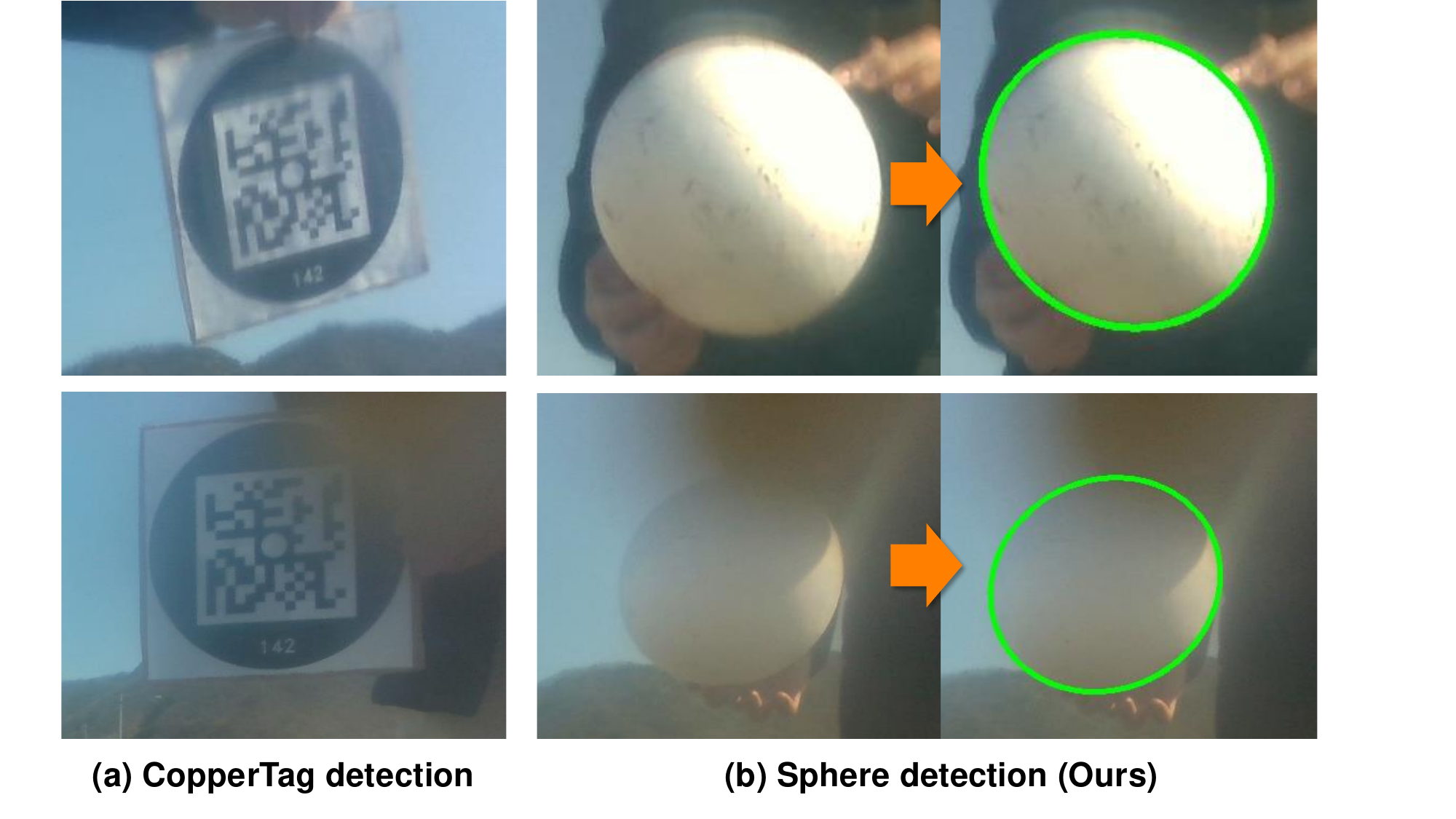}
    \vspace{-0.2cm}
     \caption{Comparison of detection result between CopperTag and spherical target under sensor contamination. (a) CopperTag failure cases under blur (top) and soil (bottom). (b) Sphere was detected even though an artifact from the blur filter was present (top) and also handled occlusion caused by mud (bottom).}
    \label{figs:blur_mud}
    \vspace{-0.4cm}
\end{figure}

Both experiments were performed in a planetary environment. As shown in \tabref{tab:detection} and \figref{figs:blur_mud}, AprilTag often failed under challenging viewing angles, lighting, and occlusions. CopperTag remained robust under moderate angle changes and mild contamination, but could not handle severe occlusions. In contrast, the spherical target consistently achieved high detection rates in all conditions.


\begin{table}[t]
\caption{Detection results under different angle, target and sensor contamination}
\label{tab:detection}
\centering\resizebox{0.48\textwidth}{!}{%
\begin{tabular}{cc|ccccc|ccccc|c}
\hline
\multicolumn{2}{c|}{}                                  & \multicolumn{5}{c|}{AprilTag}                                                                                                                                                                                                                     & \multicolumn{5}{c|}{CopperTag}                                                                                                                                                                                                                       & Sphere                                         \\ \hline
\multicolumn{2}{c|}{Degree}                            & 0°                                              & 15°                                             & 30°                                          & 45°                                             & 60°                                          & 0°                                             & 15°                                            & 30°                                            & 45°                                             & 60°                                             & 0°$\sim$60°                                    \\ \hline

\multicolumn{1}{c|}{\multirow{3}{*}{\begin{tabular}[c]{@{}c@{}}Target\\ Cont.$\dagger$\end{tabular}}}& Easy    & \multicolumn{1}{c}{\textcolor{red}{$\triangle$}} & \multicolumn{1}{c}{\textcolor{red}{$\triangle$}} & \multicolumn{1}{c}{\textcolor{red}{$\times$}} & \multicolumn{1}{c}{\textcolor{red}{$\triangle$}} & \multicolumn{1}{c|}{\textcolor{red}{$\times$}} & \multicolumn{1}{c}{\textcolor{red}{$\bigcirc$}} & \multicolumn{1}{c}{\textcolor{red}{$\bigcirc$}} & \multicolumn{1}{c}{\textcolor{red}{$\bigcirc$}} & \multicolumn{1}{c}{\textcolor{red}{$\bigcirc$}} & \multicolumn{1}{c|}{\textcolor{red}{$\bigcirc$}} & \multicolumn{1}{c}{\textcolor{red}{$\bigcirc$}} \\ 

\multicolumn{1}{c|}{} & Medium  & \multicolumn{1}{c}{\textcolor{red}{$\times$}} & \multicolumn{1}{c}{\textcolor{red}{$\times$}} & \multicolumn{1}{c}{\textcolor{red}{$\times$}} & \multicolumn{1}{c}{\textcolor{red}{$\times$}} & \multicolumn{1}{c|}{\textcolor{red}{$\times$}} & \multicolumn{1}{c}{\textcolor{red}{$\bigcirc$}} & \multicolumn{1}{c}{\textcolor{red}{$\bigcirc$}} & \multicolumn{1}{c}{\textcolor{red}{$\bigcirc$}} & \multicolumn{1}{c}{\textcolor{red}{$\bigcirc$}} & \multicolumn{1}{c|}{\textcolor{red}{$\bigcirc$}} & \multicolumn{1}{c}{\textcolor{red}{$\bigcirc$}} \\ 

\multicolumn{1}{c|}{} & Extreme & \multicolumn{1}{c}{\textcolor{red}{$\times$}} & \multicolumn{1}{c}{\textcolor{red}{$\times$}} & \multicolumn{1}{c}{\textcolor{red}{$\times$}} & \multicolumn{1}{c}{\textcolor{red}{$\times$}} & \multicolumn{1}{c|}{\textcolor{red}{$\times$}} & \multicolumn{1}{c}{\textcolor{red}{$\times$}} & \multicolumn{1}{c}{\textcolor{red}{$\times$}} & \multicolumn{1}{c}{\textcolor{red}{$\times$}} & \multicolumn{1}{c}{\textcolor{red}{$\triangle$}} & \multicolumn{1}{c|}{\textcolor{red}{$\triangle$}} & \multicolumn{1}{c}{\textcolor{red}{$\bigcirc$}} \\ 
\hline
\hline
\multicolumn{1}{c|}{\multirow{2}{*}{\begin{tabular}[c]{@{}c@{}}Sensor\\ Cont.$\dagger$\end{tabular}}} & Blur    & \multicolumn{5}{c|}{1/6}                                                                        & \multicolumn{5}{c|}{4/6}          &   \multicolumn{1}{c} {6/6}       \\
\multicolumn{1}{c|}{}                        & Mud     & \multicolumn{5}{c|}{0/6}                                                                        & \multicolumn{5}{c|}{3/6}          &   \multicolumn{1}{c} {6/6}        \\   
\hline
\multicolumn{13}{r}{ ($\dagger$) Cont. represents the contamination.} \\
\end{tabular}}
\vspace{-0.5cm}
\end{table}

\begin{table*}[!t]
\caption{Calibration Results of various sensor configurations}
\label{tab:result}

\centering\resizebox{0.90\textwidth}{!}{\scriptsize
{
\begin{tabular}{cccccccccc}
\hline 
\multicolumn{10}{c}{\textbf{Ouster OS1-32}}                                                                                                                                                     \\ \hline
\multicolumn{1}{c|}{\multirow{2}{*}{Method}} & \multicolumn{3}{c|}{Camera 1}                   & \multicolumn{3}{c|}{Camera 2}                   & \multicolumn{3}{c}{Camera 3} \\ \cline{2-10} 
\multicolumn{1}{c|}{}                        & Trans (m) & Rot (°) & \multicolumn{1}{c|}{Proj(pix)} & Trans (m) & Rot (°) & \multicolumn{1}{c|}{Proj(pix)} & Trans (m)  & Rot (°)  & Proj(pix) \\ \hline
\multicolumn{1}{c|}{\cite{toth2020automatic}*      } & 0.071     & 1.090   & \multicolumn{1}{c|}{43.009}           & 0.056     & 0.889   & \multicolumn{1}{c|}{11.899}       & 0.107      & 2.997    &  28.166   \\
\multicolumn{1}{c|}{\cite{grammatikopoulos2022effective}  } & N/A       & N/A     &    \multicolumn{1}{c|}{N/A}     & N/A       & N/A      & \multicolumn{1}{c|}{N/A}   &  N/A       & N/A      &  N/A   \\
\multicolumn{1}{c|}{Ours                                  } & \textbf{0.039}     & \textbf{0.378}   &  \multicolumn{1}{c|}{\textbf{2.142}}     & \textbf{0.018}     & \textbf{0.157}   &  \multicolumn{1}{c|}{\textbf{1.873}} & \textbf{0.028}      & \textbf{0.587}    &  \textbf{0.994}       \\ \hline 
\end{tabular}}}

\vspace{0.3cm}

\centering\resizebox{0.90\textwidth}{!}{\tiny
{
\begin{tabular}{cccccccccc}
\hline 
\multicolumn{10}{c}{\textbf{SOSLAB ML-X 120}}                                                                                                                                                     \\ \hline
\multicolumn{1}{c|}{\multirow{2}{*}{Method}} & \multicolumn{3}{c|}{Camera 1}                   & \multicolumn{3}{c|}{Camera 2}                   & \multicolumn{3}{c}{Camera 3} \\ \cline{2-10} 
\multicolumn{1}{c|}{}                        & Trans (m) & Rot (°) & \multicolumn{1}{c|}{Proj(pix)} & Trans (m) & Rot (°) & \multicolumn{1}{c|}{Proj(pix)} & Trans (m)  & Rot (°)  & Proj(pix) \\ \hline

\multicolumn{1}{c|}{\cite{toth2020automatic}*} & 0.035     & 1.502   & \multicolumn{1}{c|}{24.994}          & 0.078     & 2.055   & \multicolumn{1}{c|}{42.512}       & 0.078      & 1.800    &  25.669   \\
\multicolumn{1}{c|}{\cite{grammatikopoulos2022effective}  } & N/A       & N/A     & \multicolumn{1}{c|}{N/A}       & N/A       & N/A     & \multicolumn{1}{c|}{N/A}    & N/A        & N/A      &  N/A   \\
\multicolumn{1}{c|}{Ours                                  } & \textbf{0.016}     & \textbf{0.492}   & \multicolumn{1}{c|}{\textbf{2.142}}     & \textbf{0.026}     & \textbf{0.450}   &  \multicolumn{1}{c|}{\textbf{1.792}} & \textbf{0.010}      & \textbf{0.417}    &  \textbf{1.258}       \\ \hline 
\end{tabular}}}

\vspace{0.3cm}
\centering\resizebox{0.90\textwidth}{!}{\tiny
{
\begin{tabular}{cccccccccc}
\hline 
\multicolumn{10}{c}{\textbf{Livox MID-360}}                                                                                                                                                     \\ \hline
\multicolumn{1}{c|}{\multirow{2}{*}{Method}} & \multicolumn{3}{c|}{Camera 1}                   & \multicolumn{3}{c|}{Camera 2}                   & \multicolumn{3}{c}{Camera 3} \\ \cline{2-10} 
\multicolumn{1}{c|}{}                        & Trans (m) & Rot (°) & \multicolumn{1}{c|}{Proj(pix)} & Trans (m) & Rot (°) & \multicolumn{1}{c|}{Proj(pix)} & Trans (m)  & Rot (°)  & Proj(pix) \\ \hline

\multicolumn{1}{c|}{\cite{toth2020automatic}* } & 0.035     & 0.756   &  \multicolumn{1}{c|}{10.401}        & 0.056     & 0.889   &       \multicolumn{1}{c|}{67.578} & 0.082      & 3.234    &  29.254   \\
\multicolumn{1}{c|}{\cite{grammatikopoulos2022effective}  } & 0.164     & 1.275   &  \multicolumn{1}{c|}{2.947}   & 0.024     & 0.615   &  \multicolumn{1}{c|}{2.058} &  N/A       & N/A      &  N/A   \\
\multicolumn{1}{c|}{Ours                                  } & \textbf{0.030}     & \textbf{0.356}   &  \multicolumn{1}{c|}{\textbf{1.556}}   & \textbf{0.018}     & \textbf{0.157}   &  \multicolumn{1}{c|}{\textbf{1.644}} & \textbf{0.028}      & \textbf{0.587}    &  \textbf{1.472}       \\ \hline 
\multicolumn{10}{r}{The targetless calibration methods \cite{koide2023general, iyer2018calibnet} were excluded from the table because they did not produce valid results in the test cases.} \\
\vspace{-0.5cm}
\end{tabular}}}
\end{table*}

\subsection{Calibration test with various sensor combinations}
For the rigorous evaluation, we selected well-known methods of both target-based and targetless approaches. The target-based methods included a spherical target approach \cite{toth2020automatic} and a method that shares our goal of easy target setup, which uses retro-reflective tape and AprilTag attached to a planar board \cite{grammatikopoulos2022effective}. For the targetless methods, we compared our method against \cite{koide2023general} and~\cite{iyer2018calibnet}. However, these targetless methods did not perform consistently across all sensor configurations, likely due to difficulties in extracting and matching features in low-texture environments.






The experiment was conducted in a construction field condition with 10 scenes for each configuration using an intact spherical target mounted on a quadruped robot as depicted in \figref{figs:environments}(b). Other targets were difficult to mount on the robot, so they were acquired manually by a human. We evaluated all combinations of sensors for each target as shown in \tabref{tab:result}. The results for each setup are quantified using three metrics: translation error, rotation error, and reprojection error. The notation N/A indicates cases where results could not be obtained or where the rotation and translation errors were excessively large (exceeding 1m in translation, 10$\degree$ in rotation, and 20 pixels in reprojection). 

We modified \citet{toth2020automatic}, which is not well-suited for detecting small spheres, into \cite{toth2020automatic}* by providing a mask image with an ellipse from SAM and a cropped sphere region from the pointcloud to facilitate detection. In the method utilizing tag and reflective tape \cite{grammatikopoulos2022effective}, we encountered difficulties with the D455 camera in outdoor conditions, as sunlight introduced noise (such as a purple tint), hindering board detection. In the case of the \textsc{OS}, the intersection of the two tapes was incorrectly detected due to the excessively high intensity of the reflective tape. Additionally, for the \textsc{SOS}, the reflective tape could not be distinguished from the panel, resulting in the failure to detect key points.

Overall, our experiments confirm that the proposed method delivers superior performance compared to the other approaches. The results were consistently reliable across different LiDAR types and in the presence of camera noise.

\subsection{Calibration results on corrupted spheres}
\begin{figure}[h]
    \centering
	\def\width{0.9\columnwidth}%
	\includegraphics[width=\width, trim=0 0 12.5cm 0, clip]{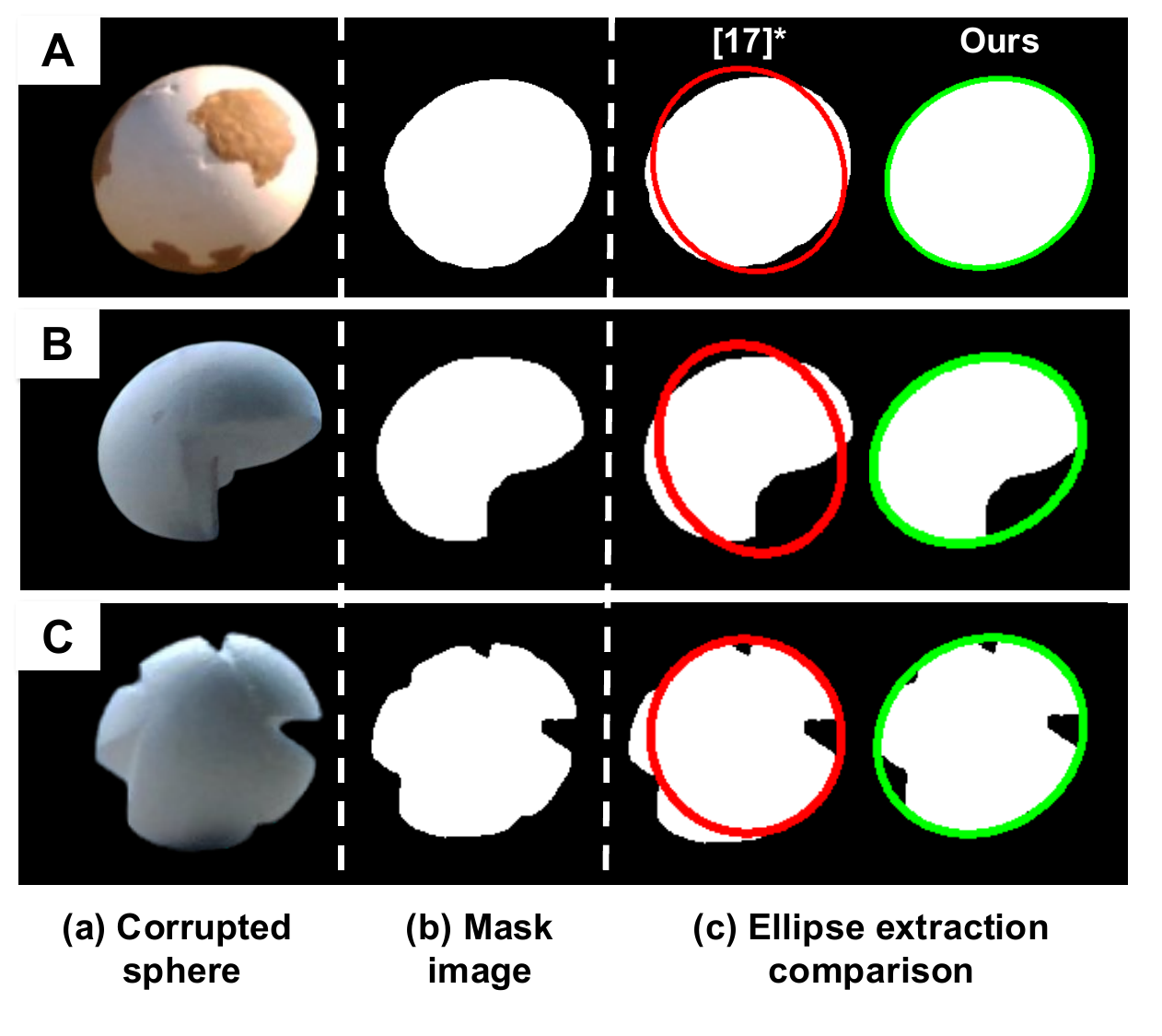}
    \vspace{-0.3cm}
     \caption{Experiments on damaged sphere targets. (a) Differently corrupted spheres. (b) Mask image from each sphere. (c) Comparison of ellipse extraction between \cite{toth2020automatic}* and ours. }
    \label{figs:3_spheres}
    \vspace{-0.3cm}
\end{figure}

We also conducted experiments to evaluate the calibration effectiveness of the proposed method in addressing both contaminated and damaged spheres, compared to \cite{toth2020automatic}*. \figref{figs:3_spheres}(a) shows a contaminated sphere along with two spheres exhibiting different types of damage: \textbf{A}, which is the extreme contamination case in \figref{figs:cont_targets}, \textbf{B}, which is truncated by approximately 25\%, and \textbf{C}, which is covered with multiple scratches. For the contamination test, the sphere was carried by the manipulator mounted on the rover in a planetary test environment as depicted in \figref{figs:environments}(a), and eight scenes were collected with \textsc{Cam2}-\textsc{OS}. Unlike the damage case, which lost data, the contamination case preserved the sphere and yielded consistent shapes across the three LiDARs. Therefore, the experiments were conducted using only one sensor pair.

\begin{table}[]
\caption{Comparison of calibration results under contamination and damage scenarios.}
\centering\resizebox{0.48\textwidth}{!}{\scriptsize
\begin{tabular}{cl|cc|cc}
\hline
                       &          & \multicolumn{2}{c|}{Ours}         & \multicolumn{2}{c}{\cite{toth2020automatic}*} \\ \cline{3-6} 
                       &          & Trans (m)       & Rot (°)         & Trans (m)  & Rot (°)          \\ \hline
\multicolumn{1}{c|}{A} & Cam2-OS  & \textbf{0.0214} & \textbf{0.4802} & 0.0366     & 0.6728           \\ \hline
\multicolumn{1}{c|}{}  & Cam1-OS  & \textbf{0.0308} & \textbf{1.2275} & 0.0402     & 1.5386           \\
\multicolumn{1}{c|}{B} & Cam1-SOS & \textbf{0.0191} & \textbf{1.2793} & 0.1027     & 1.5424           \\
\multicolumn{1}{c|}{}  & Cam1-MID & \textbf{0.0144} & \textbf{1.4118} & 0.0298     & 1.9757           \\ \hline
\multicolumn{1}{c|}{}  & Cam1-OS  & \textbf{0.0647} & 1.6923          & 0.0681     & \textbf{1.3750}  \\
\multicolumn{1}{c|}{C} & Cam1-SOS & \textbf{0.0589} & \textbf{2.8878} & 0.4689     & 5.3793           \\
\multicolumn{1}{c|}{}  & Cam1-MID & \textbf{0.0063} & \textbf{1.2548} & 0.0536     & 1.800            \\ \hline \hline
\multicolumn{2}{c|}{Average}      & \textbf{0.0308} & \textbf{1.4620} & 0.1143     & 2.0405           \\ \hline
\end{tabular}}
\vspace{-0.4cm}
\label{tab:corruption}
\end{table}

For the damage test, the sphere was mounted on a quadruped robot in a field condition, and 8 scenes were each collected with \textsc{Cam1}-\{\textsc{OS}, \textsc{SOS}, \textsc{MID}\}. In both experiments, data were acquired to ensure that the corrupted regions were prominently visible. \figref{figs:3_spheres}(b) shows soil contamination, partial truncation, and inflicted scratches degraded the mask image. The comparison of ellipse extraction is presented in \figref{figs:3_spheres}(c), which indicates that the comparison method was unable to account for these factors, resulting in incorrect ellipse detection. For quantitative evaluation, \tabref{tab:corruption} summarizes the calibration results for three corrupted targets in \figref{figs:3_spheres}(a). The proposed method demonstrates robustness and resilience to damage in most cases, and in cases where it does not, the differences are negligible. These results suggest our method can achieve robust calibration even in challenging field environments with various disturbances.

\section{CONCLUSION}

This paper presents a novel LiDAR-camera extrinsic calibration methodology that leverages a spherical target for multi-robot system in outdoor environments, where human intervention is challenging. We address potential corruption by categorizing it into two types: contamination and damage, both of which can affect the target and sensor.

Since we used a small sphere that can be carried by the robot, we employed SAM to efficiently detect it with the camera and applied a robust ellipse detection algorithm. Moreover, due to significant noise in the LiDAR data for the sphere, we devised a hierarchical weighted sum approach to accurately extract the sphere center.

Furthermore, our experiments demonstrate that, compared to other targets, the spherical target remains reliably detectable despite variations in viewing angles, as well as target and sensor contamination. We also validate the robustness of our method through experiments using LiDARs with different mechanical characteristics and cameras positioned at different viewing angles, showing its superiority over existing approaches and its resilience to target corruption.

\footnotesize	
\bibliographystyle{IEEEtranN} 
\bibliography{packages/string-short, packages/references}

\begin{thebibliography}{43}
\providecommand{\natexlab}[1]{#1}
\providecommand{\url}[1]{#1}
\csname url@samestyle\endcsname
\providecommand{\newblock}{\relax}
\providecommand{\bibinfo}[2]{#2}
\providecommand{\BIBentrySTDinterwordspacing}{\spaceskip=0pt\relax}
\providecommand{\BIBentryALTinterwordstretchfactor}{4}
\providecommand{\BIBentryALTinterwordspacing}{\spaceskip=\fontdimen2\font plus
\BIBentryALTinterwordstretchfactor\fontdimen3\font minus \fontdimen4\font\relax}
\providecommand{\BIBforeignlanguage}[2]{{%
\expandafter\ifx\csname l@#1\endcsname\relax
\typeout{** WARNING: IEEEtranN.bst: No hyphenation pattern has been}%
\typeout{** loaded for the language `#1'. Using the pattern for}%
\typeout{** the default language instead.}%
\else
\language=\csname l@#1\endcsname
\fi
#2}}
\providecommand{\BIBdecl}{\relax}
\BIBdecl

\bibitem[Yeong et~al.(2021)Yeong, Velasco-Hernandez, Barry, and Walsh]{yeong2021sensor}
D.~J. Yeong, G.~Velasco-Hernandez, J.~Barry, and J.~Walsh, ``Sensor and sensor fusion technology in autonomous vehicles: A review,'' \emph{Sensors}, vol.~21, no.~6, p. 2140, 2021.

\bibitem[Xu et~al.(2022)Xu, Zhang, Yang, Cao, Wang, Ran, Tan, and Luo]{xu2022review}
X.~Xu, L.~Zhang, J.~Yang, C.~Cao, W.~Wang, Y.~Ran, Z.~Tan, and M.~Luo, ``A review of multi-sensor fusion slam systems based on 3d lidar,'' \emph{Remote Sensing}, vol.~14, no.~12, p. 2835, 2022.

\bibitem[Tran et~al.(2023)Tran, Ahlgren, Depcik, and He]{tran2023adaptive}
D.~M. Tran, N.~Ahlgren, C.~Depcik, and H.~He, ``Adaptive active fusion of camera and single-point lidar for depth estimation,'' \emph{IEEE Transactions on Instrumentation and Measurement}, vol.~72, pp. 1--9, 2023.

\bibitem[Jing et~al.(2023)Jing, Yuan, and Hong]{jing2023online}
Y.~Jing, C.~Yuan, and X.~Hong, ``Online calibration between camera and lidar with spatial-temporal photometric consistency,'' \emph{IEEE Robotics and Automation Letters}, 2023.

\bibitem[Varadharajan and Beltrame(2025)]{varadharajan2025multi}
V.~S. Varadharajan and G.~Beltrame, ``A multi-robot exploration planner for space applications,'' \emph{IEEE Robotics and Automation Letters}, 2025.

\bibitem[Kim et~al.(2025)Kim, Choi, Kim, Yang, Cho, Lim, and Cho]{kim2025skid}
H.~Kim, J.~Choi, J.~Kim, G.~Yang, D.~Cho, H.~Lim, and Y.~Cho, ``Skid-slam: Robust, lightweight, and distributed multi-robot lidar slam in resource-constrained field environments,'' \emph{arXiv preprint arXiv:2505.08230}, 2025.

\bibitem[Ardiny and Beigzadeh(2024)]{ardiny2024enhancing}
H.~Ardiny and A.~M. Beigzadeh, ``Enhancing radioactive environment exploration with bio-inspired swarm robotics: A comparative analysis of l{\'e}vy flight and stigmergy methods,'' \emph{Robotics and Autonomous Systems}, vol. 181, p. 104794, 2024.

\bibitem[Ju et~al.(2022)Ju, Kim, Seol, and Son]{ju2022review}
C.~Ju, J.~Kim, J.~Seol, and H.~I. Son, ``A review on multirobot systems in agriculture,'' \emph{Computers and Electronics in Agriculture}, vol. 202, p. 107336, 2022.

\bibitem[Huang et~al.(2024)Huang, Zhang, Garcia, and Huang]{huang2024novel}
Z.~Huang, X.~Zhang, A.~Garcia, and X.~Huang, ``A novel, efficient and accurate method for lidar camera calibration,'' in \emph{Proc. {IEEE} Intl. Conf. on Robot. and Automat.}\hskip 1em plus 0.5em minus 0.4em\relax IEEE, 2024, pp. 14\,513--14\,519.

\bibitem[Geiger et~al.(2012)Geiger, Moosmann, Car, and Schuster]{geiger2012automatic}
A.~Geiger, F.~Moosmann, {\"O}.~Car, and B.~Schuster, ``Automatic camera and range sensor calibration using a single shot,'' in \emph{2012 IEEE international conference on robotics and automation}.\hskip 1em plus 0.5em minus 0.4em\relax IEEE, 2012, pp. 3936--3943.

\bibitem[Park et~al.(2014)Park, Yun, Won, Cho, Um, and Sim]{park2014calibration}
Y.~Park, S.~Yun, C.~S. Won, K.~Cho, K.~Um, and S.~Sim, ``Calibration between color camera and 3d lidar instruments with a polygonal planar board,'' \emph{Sensors}, vol.~14, no.~3, pp. 5333--5353, 2014.

\bibitem[Liao et~al.(2018)Liao, Chen, Liu, Wang, and Liu]{liao2018extrinsic}
Q.~Liao, Z.~Chen, Y.~Liu, Z.~Wang, and M.~Liu, ``Extrinsic calibration of lidar and camera with polygon,'' in \emph{2018 IEEE International Conference on Robotics and Biomimetics (ROBIO)}.\hskip 1em plus 0.5em minus 0.4em\relax IEEE, 2018, pp. 200--205.

\bibitem[Zhang et~al.(2023{\natexlab{a}})Zhang, Liu, Wen, Yue, Zhang, and Wang]{zhang20232}
J.~Zhang, Y.~Liu, M.~Wen, Y.~Yue, H.~Zhang, and D.~Wang, ``L 2 v 2 t 2 calib: Automatic and unified extrinsic calibration toolbox for different 3d lidar, visual camera and thermal camera,'' in \emph{2023 IEEE Intelligent Vehicles Symposium (IV)}.\hskip 1em plus 0.5em minus 0.4em\relax IEEE, 2023, pp. 1--7.

\bibitem[Beltr{\'a}n et~al.(2022)Beltr{\'a}n, Guindel, De~La~Escalera, and Garc{\'\i}a]{beltran2022automatic}
J.~Beltr{\'a}n, C.~Guindel, A.~De~La~Escalera, and F.~Garc{\'\i}a, ``Automatic extrinsic calibration method for lidar and camera sensor setups,'' \emph{IEEE Transactions on Intelligent Transportation Systems}, vol.~23, no.~10, pp. 17\,677--17\,689, 2022.

\bibitem[Wang et~al.(2017)Wang, Sakurada, and Kawaguchi]{wang2017reflectance}
W.~Wang, K.~Sakurada, and N.~Kawaguchi, ``Reflectance intensity assisted automatic and accurate extrinsic calibration of 3d lidar and panoramic camera using a printed chessboard,'' \emph{Remote Sensing}, vol.~9, no.~8, p. 851, 2017.

\bibitem[Xie et~al.(2018)Xie, Yang, Jiang, and Zhong]{xie2018pixels}
S.~Xie, D.~Yang, K.~Jiang, and Y.~Zhong, ``Pixels and 3-d points alignment method for the fusion of camera and lidar data,'' \emph{IEEE Transactions on Instrumentation and Measurement}, vol.~68, no.~10, pp. 3661--3676, 2018.

\bibitem[T{\'o}th et~al.(2020)T{\'o}th, Pusztai, and Hajder]{toth2020automatic}
T.~T{\'o}th, Z.~Pusztai, and L.~Hajder, ``Automatic lidar-camera calibration of extrinsic parameters using a spherical target,'' in \emph{Proc. {IEEE} Intl. Conf. on Robot. and Automat.}\hskip 1em plus 0.5em minus 0.4em\relax IEEE, 2020, pp. 8580--8586.

\bibitem[Zhang et~al.(2024)Zhang, Wu, Lin, Wang, and Liu]{zhang2024automatic}
G.~Zhang, K.~Wu, J.~Lin, T.~Wang, and Y.~Liu, ``Automatic extrinsic parameter calibration for camera-lidar fusion using spherical target,'' \emph{IEEE Robotics and Automation Letters}, 2024.

\bibitem[Chen et~al.(2018)Chen, Liu, Wu, Huang, Liu, Yin, Liang, Hyypp{\"a}, and Chen]{chen2018extrinsic}
S.~Chen, J.~Liu, T.~Wu, W.~Huang, K.~Liu, D.~Yin, X.~Liang, J.~Hyypp{\"a}, and R.~Chen, ``Extrinsic calibration of 2d laser rangefinders based on a mobile sphere,'' \emph{Remote Sensing}, vol.~10, no.~8, p. 1176, 2018.

\bibitem[K{\"u}mmerle et~al.(2018)K{\"u}mmerle, K{\"u}hner, and Lauer]{kummerle2018automatic}
J.~K{\"u}mmerle, T.~K{\"u}hner, and M.~Lauer, ``Automatic calibration of multiple cameras and depth sensors with a spherical target,'' in \emph{2018 IEEE/RSJ International Conference on Intelligent Robots and Systems (IROS)}.\hskip 1em plus 0.5em minus 0.4em\relax IEEE, 2018, pp. 1--8.

\bibitem[Kirillov et~al.(2023)Kirillov, Mintun, Ravi, Mao, Rolland, Gustafson, Xiao, Whitehead, Berg, Lo, et~al.]{kirillov2023segment}
A.~Kirillov, E.~Mintun, N.~Ravi, H.~Mao, C.~Rolland, L.~Gustafson, T.~Xiao, S.~Whitehead, A.~C. Berg, W.-Y. Lo \emph{et~al.}, ``Segment anything,'' in \emph{Proceedings of the IEEE/CVF International Conference on Computer Vision}, 2023, pp. 4015--4026.

\bibitem[Yuan et~al.(2021)Yuan, Liu, Hong, and Zhang]{yuan2021pixel}
C.~Yuan, X.~Liu, X.~Hong, and F.~Zhang, ``Pixel-level extrinsic self calibration of high resolution lidar and camera in targetless environments,'' \emph{IEEE Robotics and Automation Letters}, vol.~6, no.~4, pp. 7517--7524, 2021.

\bibitem[Koide et~al.(2023)Koide, Oishi, Yokozuka, and Banno]{koide2023general}
K.~Koide, S.~Oishi, M.~Yokozuka, and A.~Banno, ``General, single-shot, target-less, and automatic lidar-camera extrinsic calibration toolbox,'' in \emph{2023 IEEE International Conference on Robotics and Automation (ICRA)}.\hskip 1em plus 0.5em minus 0.4em\relax IEEE, 2023, pp. 11\,301--11\,307.

\bibitem[Ishikawa et~al.(2018)Ishikawa, Oishi, and Ikeuchi]{ishikawa2018lidar}
R.~Ishikawa, T.~Oishi, and K.~Ikeuchi, ``Lidar and camera calibration using motions estimated by sensor fusion odometry,'' in \emph{2018 IEEE/RSJ International Conference on Intelligent Robots and Systems (IROS)}.\hskip 1em plus 0.5em minus 0.4em\relax IEEE, 2018, pp. 7342--7349.

\bibitem[Schneider et~al.(2017)Schneider, Piewak, Stiller, and Franke]{schneider2017regnet}
N.~Schneider, F.~Piewak, C.~Stiller, and U.~Franke, ``Regnet: Multimodal sensor registration using deep neural networks,'' in \emph{2017 IEEE intelligent vehicles symposium (IV)}.\hskip 1em plus 0.5em minus 0.4em\relax IEEE, 2017, pp. 1803--1810.

\bibitem[Lv et~al.(2021)Lv, Wang, Dou, Ye, and Wang]{lv2021lccnet}
X.~Lv, B.~Wang, Z.~Dou, D.~Ye, and S.~Wang, ``Lccnet: Lidar and camera self-calibration using cost volume network,'' in \emph{Proceedings of the IEEE/CVF Conference on Computer Vision and Pattern Recognition}, 2021, pp. 2894--2901.

\bibitem[Claro et~al.(2023)Claro, Silva, and Pinto]{claro2023artuga}
R.~M. Claro, D.~B. Silva, and A.~M. Pinto, ``Artuga: A novel multimodal fiducial marker for aerial robotics,'' \emph{Robotics and Autonomous Systems}, vol. 163, p. 104398, 2023.

\bibitem[Fiala(2005)]{fiala2005artag}
M.~Fiala, ``Artag fiducial marker system applied to vision based spacecraft docking,'' in \emph{Proc. Intl. Conf. Intelligent Robots and Systems (IROS) 2005 Workshop on Robot Vision for Space Applications}, 2005, pp. 35--40.

\bibitem[Yahya and Arshad(2017)]{yahya2017detection}
M.~Yahya and M.~Arshad, ``Detection of markers using deep learning for docking of autonomous underwater vehicle,'' in \emph{2017 IEEE 2nd International Conference on Automatic Control and Intelligent Systems (I2CACIS)}.\hskip 1em plus 0.5em minus 0.4em\relax IEEE, 2017, pp. 179--184.

\bibitem[Bian et~al.(2024)Bian, Chen, Tian, and Ran]{bian2024coppertag}
X.~Bian, W.~Chen, X.~Tian, and D.~Ran, ``Coppertag: A real-time occlusion-resilient fiducial marker,'' in \emph{2024 IEEE International Conference on Robotics and Automation (ICRA)}.\hskip 1em plus 0.5em minus 0.4em\relax IEEE, 2024, pp. 7273--7279.

\bibitem[Chen et~al.(2021)Chen, Liu, and Xiong]{chen2021automatic}
B.~Chen, Y.~Liu, and C.~Xiong, ``Automatic checkerboard detection for robust camera calibration,'' in \emph{2021 IEEE International Conference on Multimedia and Expo (ICME)}.\hskip 1em plus 0.5em minus 0.4em\relax IEEE, 2021, pp. 1--6.

\bibitem[Uricar et~al.(2021)Uricar, Sistu, Rashed, Vobecky, Kumar, Krizek, Burger, and Yogamani]{uricar2021let}
M.~Uricar, G.~Sistu, H.~Rashed, A.~Vobecky, V.~R. Kumar, P.~Krizek, F.~Burger, and S.~Yogamani, ``Let's get dirty: Gan based data augmentation for camera lens soiling detection in autonomous driving,'' in \emph{Proceedings of the IEEE/CVF winter conference on applications of computer vision}, 2021, pp. 766--775.

\bibitem[Zhang et~al.(2023{\natexlab{b}})Zhang, Carballo, Yang, and Takeda]{zhang2023perception}
Y.~Zhang, A.~Carballo, H.~Yang, and K.~Takeda, ``Perception and sensing for autonomous vehicles under adverse weather conditions: A survey,'' \emph{ISPRS Journal of Photogrammetry and Remote Sensing}, vol. 196, pp. 146--177, 2023.

\bibitem[Canny(1986)]{canny1986computational}
J.~Canny, ``A computational approach to edge detection,'' \emph{IEEE Transactions on pattern analysis and machine intelligence}, no.~6, pp. 679--698, 1986.

\bibitem[Ranganathan(2004)]{ranganathan2004levenberg}
A.~Ranganathan, ``The levenberg-marquardt algorithm,'' \emph{Tutoral on LM algorithm}, vol.~11, no.~1, pp. 101--110, 2004.

\bibitem[T{\'o}th and Hajder(2023)]{toth2023minimal}
T.~T{\'o}th and L.~Hajder, ``A minimal solution for image-based sphere estimation,'' \emph{Proc. {IEEE} Intl. Conf. on Robot. and Automat.}, vol. 131, no.~6, pp. 1428--1447, 2023.

\bibitem[Hough(1962)]{hough1962method}
P.~V. Hough, ``Method and means for recognizing complex patterns,'' Dec.~18 1962, uS Patent 3,069,654.

\bibitem[Rusu and Cousins(2011)]{rusu20113d}
R.~B. Rusu and S.~Cousins, ``3d is here: Point cloud library (pcl),'' in \emph{2011 IEEE international conference on robotics and automation}.\hskip 1em plus 0.5em minus 0.4em\relax IEEE, 2011, pp. 1--4.

\bibitem[Laconte et~al.(2019)Laconte, Desch{\^e}nes, Labussi{\`e}re, and Pomerleau]{laconte2019lidar}
J.~Laconte, S.-P. Desch{\^e}nes, M.~Labussi{\`e}re, and F.~Pomerleau, ``Lidar measurement bias estimation via return waveform modelling in a context of 3d mapping,'' in \emph{2019 International Conference on Robotics and Automation (ICRA)}.\hskip 1em plus 0.5em minus 0.4em\relax IEEE, 2019, pp. 8100--8106.

\bibitem[Zhou et~al.(2018)Zhou, Li, and Kaess]{zhou2018automatic}
L.~Zhou, Z.~Li, and M.~Kaess, ``Automatic extrinsic calibration of a camera and a 3d lidar using line and plane correspondences,'' in \emph{2018 IEEE/RSJ International Conference on Intelligent Robots and Systems (IROS)}.\hskip 1em plus 0.5em minus 0.4em\relax IEEE, 2018, pp. 5562--5569.

\bibitem[Olson(2011)]{olson2011apriltag}
E.~Olson, ``Apriltag: A robust and flexible visual fiducial system,'' in \emph{2011 IEEE international conference on robotics and automation}.\hskip 1em plus 0.5em minus 0.4em\relax IEEE, 2011, pp. 3400--3407.

\bibitem[Grammatikopoulos et~al.(2022)Grammatikopoulos, Papanagnou, Venianakis, Kalisperakis, and Stentoumis]{grammatikopoulos2022effective}
L.~Grammatikopoulos, A.~Papanagnou, A.~Venianakis, I.~Kalisperakis, and C.~Stentoumis, ``An effective camera-to-lidar spatiotemporal calibration based on a simple calibration target,'' \emph{Sensors}, vol.~22, no.~15, p. 5576, 2022.

\bibitem[Iyer et~al.(2018)Iyer, Ram, Murthy, and Krishna]{iyer2018calibnet}
G.~Iyer, R.~K. Ram, J.~K. Murthy, and K.~M. Krishna, ``Calibnet: Geometrically supervised extrinsic calibration using 3d spatial transformer networks,'' in \emph{2018 IEEE/RSJ International Conference on Intelligent Robots and Systems (IROS)}.\hskip 1em plus 0.5em minus 0.4em\relax IEEE, 2018, pp. 1110--1117.

\end{thebibliography}

\end{document}